\documentclass[sigconf]{acmart}   
\renewcommand\footnotetextcopyrightpermission[1]{} 

\usepackage{amsmath,amsfonts,bm}

\def\eqref#1{equation~\ref{#1}}

\def\1{\bm{1}}

\def\vf{{\bm{f}}}
\def\vg{{\bm{g}}}

\def\vt{{\bm{t}}}

\def\vv{{\bm{v}}}

\DeclareMathAlphabet{\mathsfit}{\encodingdefault}{\sfdefault}{m}{sl}
\SetMathAlphabet{\mathsfit}{bold}{\encodingdefault}{\sfdefault}{bx}{n}

\newcommand{\softmax}{\mathrm{softmax}}

\usepackage{caption}
\usepackage{subcaption}

\DeclareMathOperator{\MLP}{MLP}

\AtBeginDocument{%
  \providecommand\BibTeX{{%
    \normalfont B\kern-0.5em{\scshape i\kern-0.25em b}\kern-0.8em\TeX}}}

\begin{document}

\title{Personalized Fashion Recommendation from Personal Social Media Data: An Item-to-Set Metric Learning Approach}



\author{Haitian Zheng}
\affiliation{
 \institution{University of Rochester}
 \city{Rochester}
 \state{NY}
 \country{USA}
 }
 \email{hzheng15@ur.rochester.edu}

\author{Kefei Wu}
\affiliation{
 \institution{University of Rochester}
 \city{Rochester}
 \state{NY}
 \country{USA}
 }
 \email{kwu19@ur.rochester.edu}

\author{Jong-Hwi Park}
\affiliation{
 \institution{University of Rochester}
 \city{Rochester}
 \state{NY}
 \country{USA}
 }
 \email{jpark177@ur.rochester.edu}

\author{Wei Zhu}
\affiliation{
 \institution{University of Rochester}
 \city{Rochester}
 \state{NY}
 \country{USA}
 }
 \email{wzhu15@ur.rochester.edu}

\author{Jiebo Luo}
\affiliation{
 \institution{University of Rochester}
 \city{Rochester}
 \state{NY}
 \country{USA}
 }
 \email{jluo@cs.rochester.edu}

\newcommand{\best}[1]{{\bf{#1}}}

\begin{abstract}
With the growth of online shopping for fashion products, accurate fashion recommendation has become a critical problem. 
Meanwhile, social networks provide an open and new data source for personalized fashion analysis.
In this work, we study the problem of personalized fashion recommendation from social media data, i.e. recommending new outfits to social media users that fit their fashion preferences.
To this end, we present an item-to-set metric learning framework that learns to compute the similarity between a set of historical fashion items of a user to a new fashion item.
To extract features from multi-modal street-view fashion items, we propose an embedding module that performs multi-modality feature extraction and cross-modality gated fusion. 
To validate the effectiveness of our approach, we collect a real-world social media dataset.
Extensive experiments on the collected dataset show the superior performance of our proposed approach.
\end{abstract}

\ccsdesc[500]{Human-centered computing~Social networking sites}
\ccsdesc[500]{Applied computing~Online shopping}

\keywords{Fashion recommendation, Recommendation system, Social Media, Metric learning}

\maketitle


\section{Introduction}

\begin{figure}[t!]
  \centering
  \includegraphics[width=\linewidth]{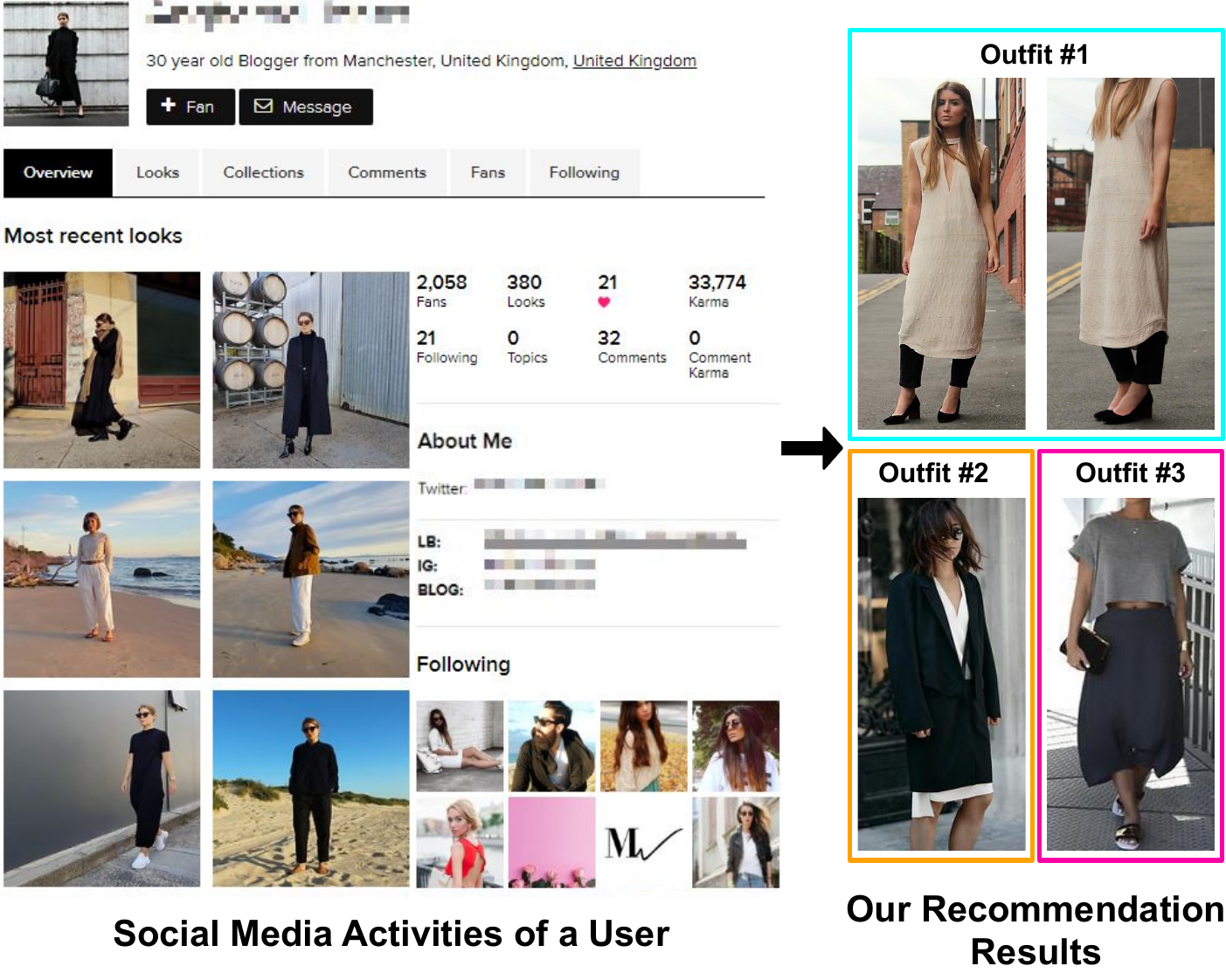}
  \caption{An example user web page on lookbook.nu. A user web page contains street-view fashion selfie posts uploaded by the user, which reveal his/her personal fashion interests. Taking such user activities as inputs, our model can recommend personalized outfits based on their fashion preferences. 
    In this specific example, the top-3 recommendations in the figure reflect minimal and monotonic color looks that the user prefers. Note that outfit \#1 correctly predicts one of her own outfits in the testing pool.} 
  \label{fig:teaser}
\end{figure}

With the thriving social networks, people start to share everyday moments online. For instance, they share the place they visited, the food they had, and the outfit they wore. There are multiple fashion-oriented online communities where users show off their dressing styles and connect to new people that share similar fashion interests.
As an example, Lookbook\footnote{https://lookbook.nu/} users can showcase their fashion styles with various street-view selfie posts (Fig.~\ref{fig:teaser}), which no doubt reveal their individual fashion preferences.
This emerging trend presents a new opportunity for personalized fashion analysis through analyzing the user-created contents, and allows us to uncover the fashion interest of individual users at personal levels.


In this work, we study the problem of personalized fashion recommendation with personal social media data, which seeks to recommend new fashion outfits based on online activities of social network users.
Although there have been a number of studies on clothing retrieval and recommendation~\cite{hadi2015buy,huang2015cross,kuang2019fashion,wang2017clothing,jagadeesh2014large,simo2015neuroaesthetics,iwata2011fashion,hu2015collaborative,liu2017deepstyle,ma2019and}, exploiting personal social media data for fashion recommendation is fundamentally challenging and less explored.
In particular, the online activities of social media users are often no more than street-view selfie with additional word descriptions. The granularity of such data is much coarser than other explored data types, such as transaction records~\cite{cardoso2018product}, human evaluation~\cite{tangseng2017recommending,tangseng2020toward}, garment item annotations~\cite{tangseng2017recommending,li2017mining,hu2015collaborative,yu2018aesthetic} and annotated attributes~\cite{yang2019interpretable}. 
As a result, most established models are not directly applicable to our task due to the lack of supervision.

Without requiring other fine-grained supervision beyond the selfie posts, we propose a self-supervised approach for effective and personalized fashion recommendation
We regard the selfie posts of users as either \emph{a set} that reveals their personal fashion preferences, or outfit \emph{items} of to-be-recommended items.
Upon this basis, we propose to learn an \emph{item-to-set} metric that measures the similarities between a set and items for personalized recommendation.
To this end, we propose a self-supervised task that seeks to minimize the item-to-set distance for the set and items of a user, while maximizing such distances for sets and items of different users. Benefiting from such a training scheme, our framework is able to perform personalized recommendation without requiring any additional supervision such as transaction records~\cite{cardoso2018product} or human evaluation~\cite{tangseng2017recommending,tangseng2020toward}.

Although metric learning has been well-studied in the literature~\cite{contrastive,triplet,npair,moco}, learning such an item-to-set metric is previously unexplored and faces new challenges.
In reality, a user can have interest in more than one fashion style. 
Therefore, the item-to-set similarity cannot be captured by an oversimplified average of multiple item-wise similarities.
Alternatively, a nearest-neighbor item-to-set metric is difficult to learn as it is susceptible to noise and outliers.

To addresses the above issues, we propose a new and generalized item-to-set metric. 
Specifically, we propose an importance weight for each item in the set.
The importance weight changes according to a different set and query, and it serves to filter out outliers and unrelated items from the set. Different from nearest-neighbor classification, it can update features of all set items and enable effective learning.
We consider two principles, namely neighboring importance and intra-set importance to implement the importance function. The neighboring importance serves to filter out set items that are far away from the new item, while the intra-set importance serves to filter the noise and outlier inside the set.

We further propose a user-specific item-to-set metric.
The new metric is motivated by the fact that different users focus on different aspects of fashion products.
As a result, the similarity metric should depend on the set of selfie posts to facilitate more targeted fashion recommendation.
To utilize user-specific information, we propose a space transform operation that transforms item features into  user-specific space before the similarity computation. The user-specific design further boosts the fashion recommendation performances.

Extracting fashion preferences information from user selfie posts involves understanding the raw fashion images and the associated text descriptions, as well as fusing information from multiple sources for better feature integration.
To this end, we design a multi-modal embedding module.
In particular, we design an image embedding module that extracts high-level fashion feature from raw selfie images. 
We also design hashtag and title embedding modules that utilize attentive averaging to extract semantic features from sets of word descriptions.
To alleviate the influence of incorrect parsing, missing modalities or typos, we design a cross-gated fusion module that performs progressive feature fusion for each modality.

To validate the effectiveness of our proposed approach, we collect a real-world social media dataset. Through extensive experiments on the network design, we validate the effectiveness of our approach. 

We highlight our contributions as follows:
\begin{itemize}
    \item We present a fashion recommendation system built on personal social media data. Our system recommends personalized outfits using a few unconstrained street-view selfie posts of users.
    \item We propose a self-supervised scheme to enable the training of the system. Our approach is based on a novel item-to-set metric learning framework that requires only the user selfie posts as the supervision.
    \item We design a multi-modal embedding module that better fuses the social media data for extraction of fashion features.
    \item We evaluate our approach on our collected social media dataset. Extensive experiments on the real world dataset demonstrates the effectiveness of our approach.
\end{itemize}




\section{Related Works}
Fashion analysis has drawn broad interest in the multimedia community. 
The recent studies on fashion analysis can be categorized into four aspects, namely: 1) fashion annotation, 2) fashion retrieval, 3) fashion composition, and 4) fashion recommendation. 

\subsection{Fashion Annotation}
Fashion annotation aims at generating fashion attributes to facilitate automatic fashion analysis. It includes clothing parsing, recognition, attributes annotation and landmark detection. 
Clothing parsing~\cite{yamaguchi2012parsing,dong2014towards,liu2016deepfashion} predicts garment items at pixel-level. Recent works in this field~\cite{liang2018look,zhou2018adaptive,gong2018instance} apply techniques from semantic segmentation~\cite{ren2015faster} and achieve significant improvements. Attribute annotation~\cite{liu2012hi,liu2012street,chen2015deep,chen2012describing} aims to generate fashion attributes from clothing images. Liu et al.~\cite{liu2016deepfashion} propose a large-scale fashion dataset with attribute annotation.
Kenan E et al.~\cite{ak2018learning} utilize weakly supervised learning to annotate attributes with localization.
Towards fashion landmark detection, Liu et al.~\cite{liu2016fashion} propose a cascading multiple convolutional network to detect landmarks.
Yan et al.~\cite{yan2017unconstrained} propose a recurrent transformer network for unconstrained fashion landmark detection.

\subsection{Fashion Retrieval}
Clothing retrieval~\cite{yamaguchi2014retrieving,yamaguchi2013paper,wang2013personal} attempts to find similar clothing from a person image query. 
Typical approaches on fashion retrieval~\cite{liu2012street,wang2013personal} utilize attributes to learn a fashion representation. 
Recently, Kiapour et al.~\cite{hadi2015buy} propose a deep metric network to retrieve garment items.
Huang et al.~\cite{huang2015cross} design an attribute-aware ranking network for retrieval feature learning.
Kuang et al.~\cite{kuang2019fashion} design a graph reasoning network that learns visual similarity for fashion retrieval.
Wang et al.~\cite{wang2017clothing} design a self-learning model that learns to retrieve from image inputs.

\subsection{Fashion Composition}
Fashion composition focus on measuring whether clothing items are compatible  and aims to generate visually complementary combination of fashion items. 
To this end, Li et al.~\cite{li2017mining} propose a learning-based approach on set data for mining outfit compositions.
Han et al.~\cite{han2017learning} predict compatibility relationships of fashion items with sequence models. \cite{song2017neurostylist} learn compatibility models by Bayesian personalize ranking.
Hsiao et al. \cite{hsiao2018creating} study the problem of automatic capsule creation. 
Ma et al.~\cite{ma2019and} derive fashion knowledge using a social media database.

\subsection{Fashion Recommendation}
There are several attempts at personalized fashion recommendation.
Jagadeesh et al.~\cite{jagadeesh2014large} design a data driven model that performs complementary fashion recommendation from visual input.
Simo-Serra et al.~\cite{simo2015neuroaesthetics} propose a random field model that jointly reasons about fashionability factors of users for fashion outfit recommendation.
Iwata et al.~\cite{iwata2011fashion} propose a probabilistic topic model for
learning fashion coordinates.
Hu et al.~\cite{hu2015collaborative} propose a tensor factorization approach for collaborative fashion recommendation.
Liu et al.~\cite{liu2017deepstyle} design a visual-based model that learns style feature of items for sensing preferences of users.
Hidayati et al.~\cite{ma2019and} study the problem of fashion recommendation for personal body types.
\section{Data Construction}
\label{sec:data}
We crawl a social media dataset from a popular fashion-focused website \textit{Lookbook.nu}, where users can freely post their outfits and selfies.
The left part of Fig~\ref{fig:teaser} shows the profile of a user, with recent selfies posted by users.
We crawl a total of 2,293 personal profiles from users. For each user, we keep their 100 most recent selfie posts with their corresponding photo titles and hashtags.
The 2,293 users do not include any users with more than 7,000 fans because the latter are most likely  fashion brand’s commercial accounts which contain diverse  photos of different fitting models.

\subsection{Data Overview}
\begin{figure}[http!]
  \centering
  \includegraphics[width=\linewidth]{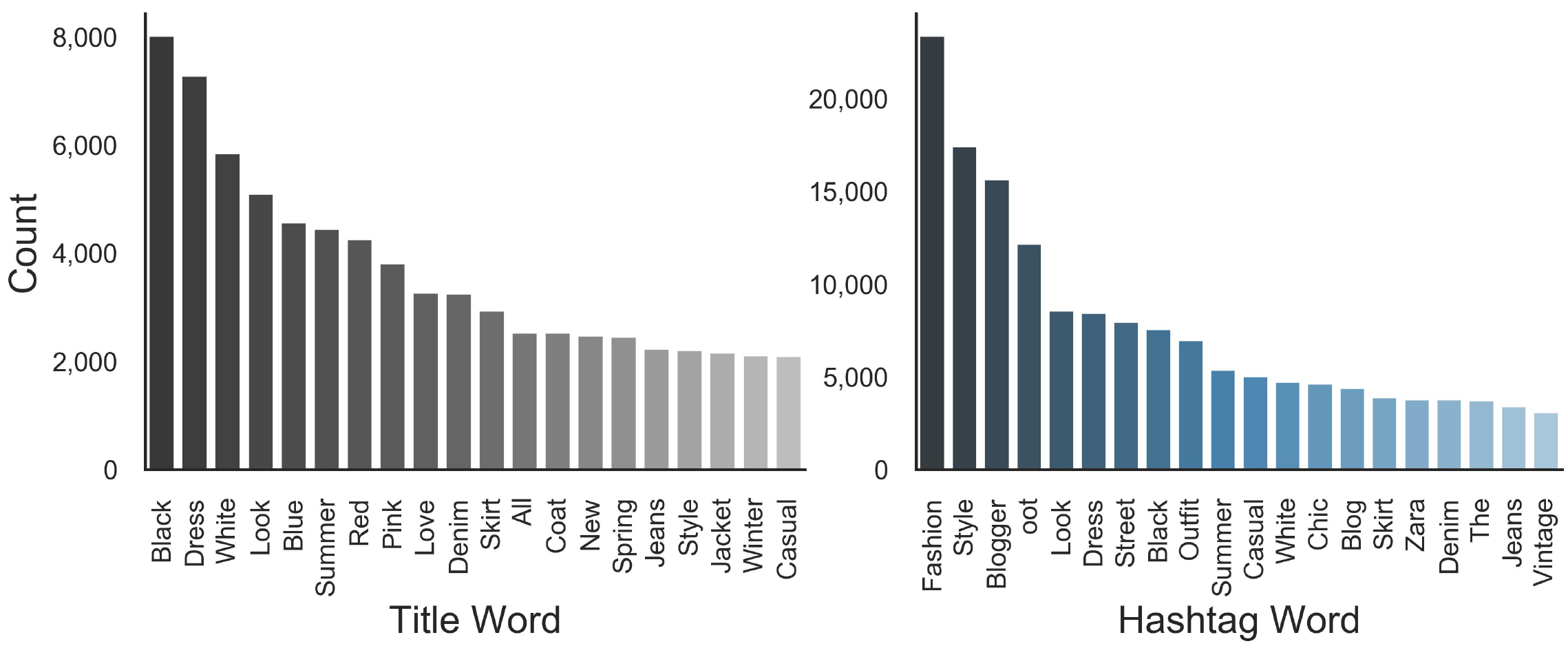}
  \vspace{-5mm}
  \caption{The word frequency for the hashtag and title modalities. Words are sorted by the frequency in the modalities.
  }
  \label{fig:item distribution graph}
\end{figure}

In Table~\ref{tab:dataset}, we show the basic statistics of our collected data, which includes user attributes such as age, number of looks, likes per picture, number of fans, and number of followings.
We also visualize the most frequent 20 words from the hashtags and titles as shown in  Fig. ~\ref{fig:item distribution graph}. The frequency plot suggests that hashtags are often words that describe the styles of outfits, such as street, blogger and summer.
On the other hand, title words are more specific and usually describe attributes and colors of outfits.
In Table~\ref{tab:dataset2}, we show garment statistics of our processed dataset. Garment statistics shows that basics garments such as upper clothes, pants and shoes have higher proportion, while accessories have lower proportion.


\begin{table}[http!]
\caption{The user statistics of our collected dataset.}
\vspace{-3mm}
\centering
\scalebox{0.9}{
    \begin{tabular}{ c|c|c|c|c|c|c }
    \toprule
    \hline
    Distribution & Min&Max&Mean&Median&Std&Skewness \\
    \hline
    Age&19 &70 &29.20 &29 &4.48 &1.88\\
    \#Looks &100 &2,414 &203.33 &158 &149.64 &5.20\\
    \#Likes &0 &8,730 &113.23 &80 &119.57 &4.23\\
    \#Fans&3 &6,672 &1,171.36 &765 &1,157.77 &1.45\\
    \#Followings &0 &21,423 &269.28 &73 &982.48 &12.87\\
    \hline
    \bottomrule
    \end{tabular}
    \label{tab:dataset}
}
\end{table}

\vspace{-3mm}


\vspace{-1mm}
\begin{table}[http!]
\caption{The garment statistics of our processed dataset. 
}
\vspace{-3mm}
\centering
\scalebox{0.9}{
    \begin{tabular}{ c | c  c || c | c c}
    \toprule
    \hline
    Garment & Count  & Proportion &Garment &Count  & Proportion \\
    \hline
    \ Upper Clothes &178,041 &.21 &Hat &49,887 &.06\\
    \ Pants &167,697 &.2 &Socks &38,627 &.05\\
    \ Shoes &156,880 &.19 &Glove &11,785 &.01\\
    \ Coat &104,710 &.12 &Scarf &9,775 &.01\\
    \ Dress &62,627 &.07 &Jumpsuits &4,619 &.01\\
    \ Skirt &60,315 &.07 & & &\\

    \hline
    \bottomrule
    \end{tabular}
    \label{tab:dataset2}
}
\end{table}

\vspace{-5mm}
\subsection{Data Prepossessing}
\subsubsection{Image Data Preprocessing}
Since the raw selfie images often contain multiple concatenated pictures, we utilize a detection model~\cite{yolo_v3} to crop person bounding-boxes, and then select the ones with the highest scores to obtain the best-fitted person images.
We also exclude grayscale images which typically cannot fully reflect the outfit styles. 

\subsubsection{Word Data Preprocessing}
For title and hashtag features, we utilize the Wikipedia pretrained GloVe text embedding to extract features. Specifically, for hashtag, we first apply the Viterbi algorithm to compute word segments. The embedding of each word is taken as input to generate hashtag features. Likewise, we use the embedding of every word from the title to generate the title features. After extracting both word and image features, features are simply concatenated as a row vector that corresponds to a single post.

\section{Method}
\label{sec:method}
As illustrated in Fig.~\ref{fig:teaser},
we consider the problem of personalized fashion recommendation from personal social media, which aims at recommending new outfits to a user based on several selfies of that user.
We intend to achieve the following objectives:
\begin{enumerate}
    \item {\it Multi-modality feature extraction.} The multi-modality social-media activities can reveal fashion preferences of individuals. 
    Our system should extract the fashion preferences of individuals based on user activities, and then match the preference with candidate outfits.
    \item {\it Multi-interest awareness.} User selfie posts can reveal multiple fashion interests that a user may have.
    Our system should effectively represent the multiple interests of a user for better recommendation.
    \item {\it User uniqueness.} Different users may pay attention to specific fashion components while being less sensitive to others.
    Our recommendation should take into account such user-specific fashion interests for a more targeted recommendation.
\end{enumerate}

To meet those objectives, we propose an embedding network in Sec.~\ref{sec:embedding} that embeds the unique activities and outfit candidates into a feature space. Next, in Sec.~\ref{sec:metric_learning}, we present an item-to-set metric learning framework that learns to match user selfie posts to new outfits. Finally, in Sec.~\ref{sec:loss}, we provide the training scheme and objective of our model.

\vspace{-3mm}
\subsection{Fashion Item Embedding Module}
\label{sec:embedding}
\begin{figure}[tp!]
  \centering
  \includegraphics[width=\linewidth]{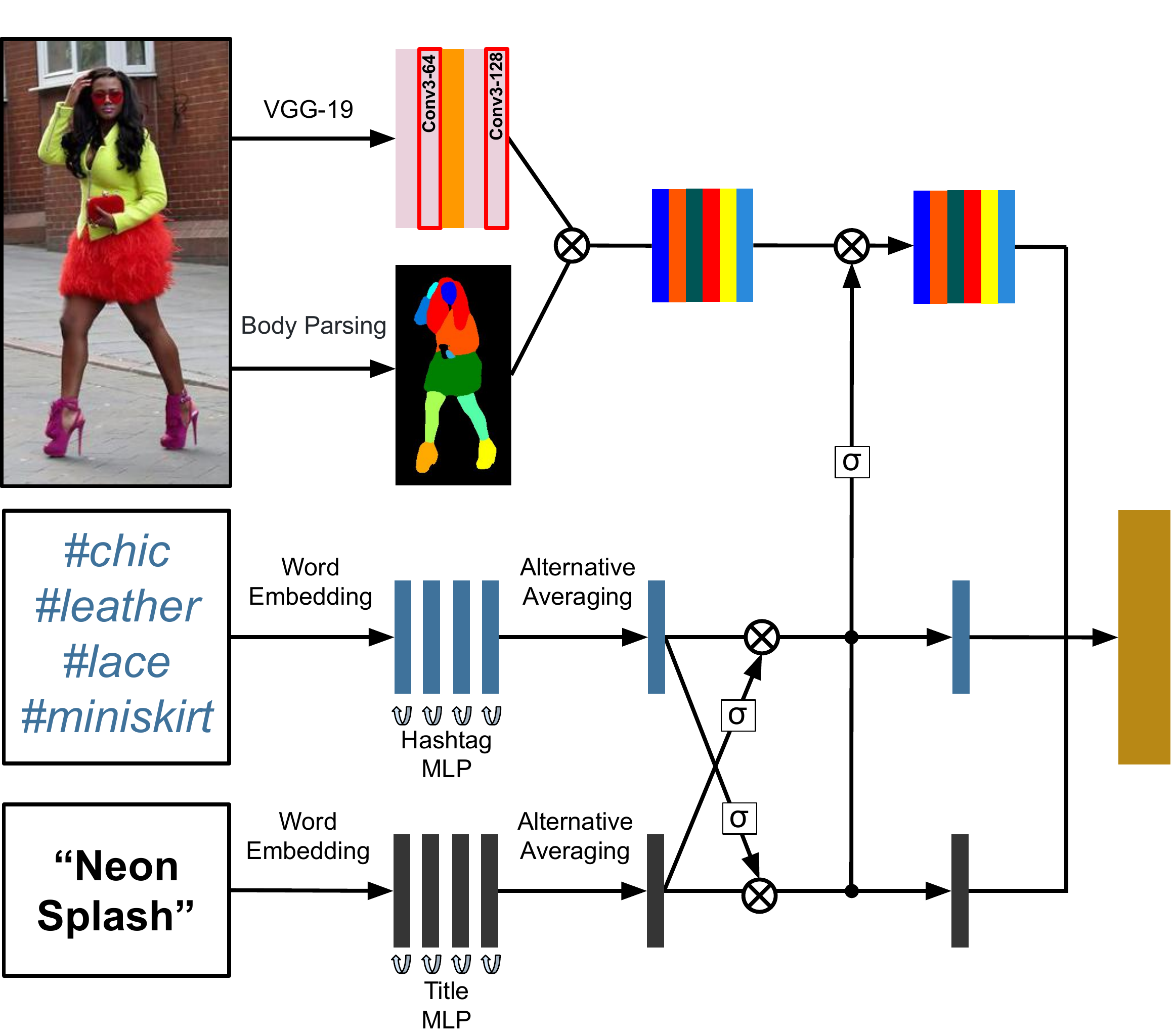}
  \vspace{-3mm}
  \caption{Our embedding module takes a multi-modal user post of as inputs and outputs a  joint embedding. We extract visual features for individual garment parts to better capture fashion clues.}
  \label{fig:embedding}
\end{figure}

Social media users often post fashion selfies with photo titles and hashtags. Such multi-modal user activities often reveal the personal fashion preferences of the users. 
To extract fashion information from user activities, we propose a fashion item embedding module. As shown in Fig.~\ref{fig:embedding}, the embedding module extracts fashion information from image-hashtag-title triplets $x=(x^{(im)},x^{(h)},x^{(t)})$ by first extracting features from the three modalities and then performing multi-modality fusion.

\subsubsection{Image Feature Extraction}
Image inputs $x_{im}$ can reveal the outfit combination and appearance preference of a user. To extract fashion information from images, we incorporate a body parsing model~\cite{liang2018look} to extract $11$ garment regions from the input image. The garment regions that we use include common garments semantics such as \texttt{\{dress}, \texttt{coat}, \texttt{pant}, \texttt{skirt\}} and exclude non-garment semantics (\texttt{face}, \texttt{hair}, \texttt{background}) or rare regions (\texttt{socks}, \texttt{sunglasses}, etc.). 
We then utilize a pre-trained image recognition model~\cite{vgg} to extract visual information from each garment regions. Specifically, we
compute the feature response of~\cite{vgg} at layer \texttt{conv1} and \texttt{conv2}, and average the feature response for each garment region. We then concatenate features of all the garment regions, resulting a $2112$-d image feature that represents the visual style of a fashion outfit. An illustrative example of such a process is shown on the top of Fig.~\ref{fig:embedding}.

\subsubsection{Hashtag Feature Extraction}
A hashtag modality input $x^{(h)}$ is a set of word embedding vectors $x^{(h)}=\{{\vv}^{(h)}_1,\cdots, {\vv}^{(h)}_m\}$ that are extracted from the hashtags of a fashion item.
In the case that the hashtag modality is missing, a zero vector is used to represent an empty hashtag set. Since semantically different words can sometimes refer to the same fashion style, e.g., corset, leatherjacket and black can refer to the goth style,
we employ an additional MLP to transform the general word embedding ${\vv}^{(h)}_{i}$ into a fashion-related embedding ${\vf}^{(h)}_{i}$.

Similar to image feature extraction, we aim at generating a fixed-length vector to represent the hashtag features. Because the size of embedding features may vary, we present an attentive averaging operation to weighted average the embedding features. Specifically, an MLP is applied to generate unnormalized weights for each feature ${\vf}^{(h)}_i$, followed by a softmax operation and weighted averaging to produce an averaged feature:
\begin{align}
\begin{aligned}
    & e_i = \MLP_2({\vf}^{(h)}_i),\\
    &\alpha_i = \frac{\exp(e_i)} {\sum_j {  \exp(e_j)   }}\\
    &\vg^{(h)} = \sum_{i=1}^{m} {\alpha_i {\vf}^{(h)}_{i}}.
\end{aligned}
\end{align}
The averaged feature $\vg^{(h)}$ serves to represent the feature of the hashtag modality. 

\subsubsection{Title Feature Extraction}
A title modality input $x^{(t)}$ is a set of word embedding vectors $x^{(t)}=\{{\vv}^{(t)}_1,\cdots, {\vv}^{(t)}_m\}$ that are extracted from the title words. In a similar fashion to the hashtag feature extraction, we utilize an MLP to extract fashion-aware features. Next, attentive averaging is used to extract features for title modality.

\subsubsection{Cross-modality Gated Fusion}
The multi-modality features extracted in the previous steps often suffer from incorrect body parsing, missing modalities or misspelling.
To improve the quality of features, we propose a multi-modality cross-gating scheme that sequentially integrates  information from alternative modalities for feature fusion.


Specifically, since title and hashtag features are less noisy than image features and they carry complementary semantics information, our scheme first performs a cross-gating operation between hashtag and title modalities. The hashtag and title features are updated simultaneously using the following updating rules:
\begin{align}
\begin{aligned}
    \vf^{h} \gets \vf^{h} \odot \sigma (\MLP_{\vf\_h}(\vf^{t})), \\
    \vf^{t} \gets \vf^{t} \odot \sigma (\MLP_{\vf\_t}(\vf^{h})),
\end{aligned}
\end{align}
where $\MLP_{\vf\_h}$ and $\MLP_{\vf\_t}$ generate a filtering score in each feature dimension, respectively. The operation $\odot$ represents an element-wise product and $\sigma$ represents a sigmoid function.

In a similar fashion, the cross-filtered features from hashtag and title modalities are used to filter the low-level image features. Specifically, the image modality feature is updated by:
\begin{align}
\begin{aligned}
    f^{i} \gets \vf^{i} \odot \sigma (\MLP_{\vf\_i}([\vf^{h},\vf^{t}])), 
\end{aligned}
\end{align}
where $\MLP_{\vf\_i}$ generates the filtering score for every dimension of the image modality.

Finally, a 2-layer MLP is employed on the concatenation of features from all modalities to generate the fused feature from the selfie posts of each user:
$$\vf_i = \MLP_{fusion}([\vf^{i},\vf^{h},\vf^{t}]).$$

\begin{figure}[tp!]
  \centering
  \includegraphics[width=\linewidth]{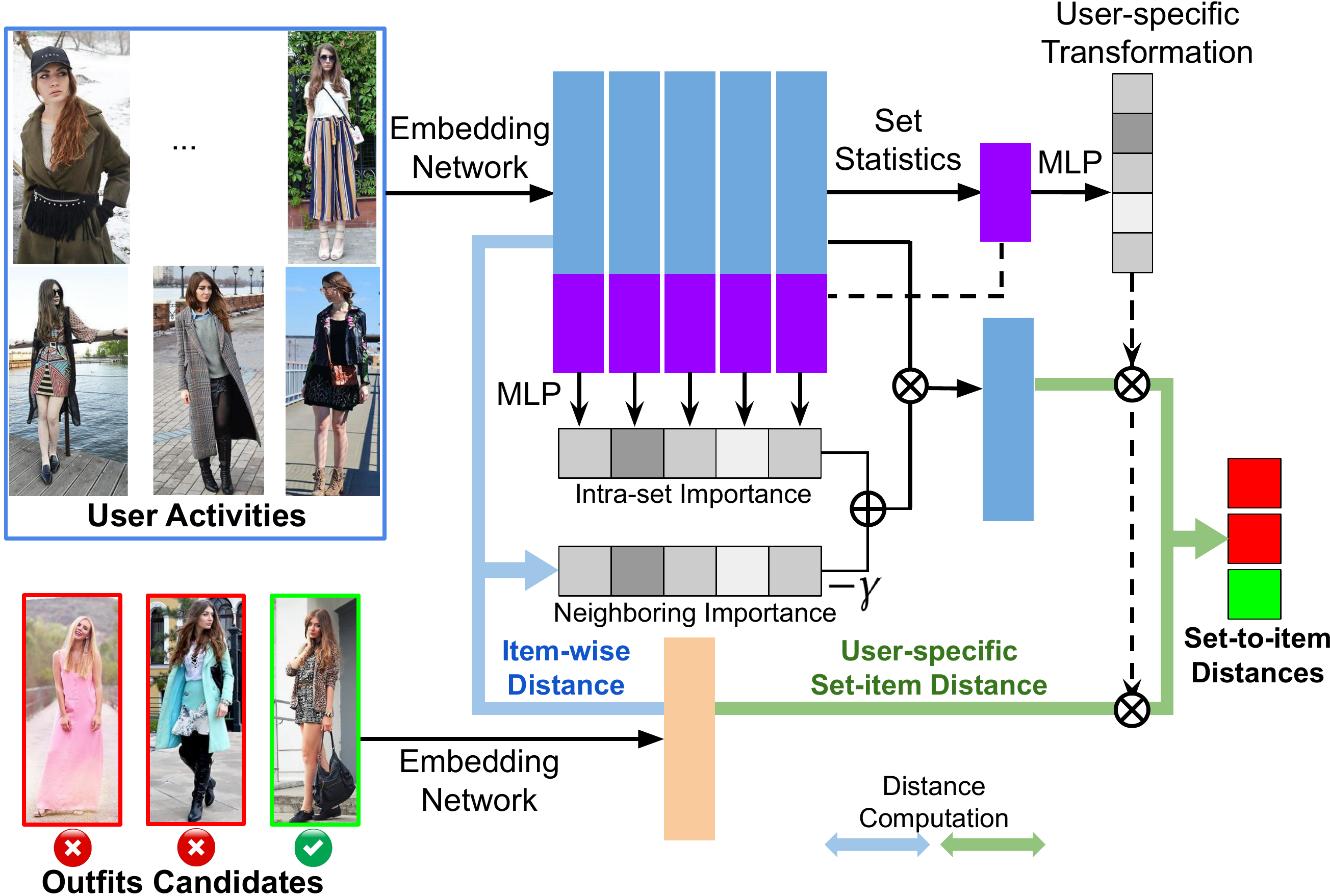}
  \vspace{-3mm}
  \caption{The pipeline of item-to-set metric learning. Given a set of activities of a user, item-to-set metric learning aims at generating low distance scores for the {\color{green}{positive item}} and high distance scores for the {\color{red}{negative items}}.
  Our approach generates both \emph{intra-set} and \emph{neighboring} importance scores to dynamically select items from the set for distance computation. Furthermore, we propose a user-specific transformation for learning user-specific metric.
  }
  \label{fig:metric_learning}
\end{figure}

\vspace{-3mm}
\subsection{Item-to-set Metric Learning}
\label{sec:metric_learning}
We consider a generalized metric learning problem that learns a similarly measurement from a set of user selfie posts $S=\{\vf_1, \cdots, \vf_k\}$ to a candidate outfit item $f$ that is in the recommendation pool. 
In the following, we will review the concept of metric learning, then propose our item-to-set metric learning framework.


\subsubsection{Item-to-set Similarity Metric}
Metric learning typically aims to learn a similarity measurement between two items. Typical metric learning approaches~\cite{contrastive,triplet,jing2020self} often regard two items $i$ and $j$ as feature points $\vf_i$ and $\vf_j$ in a normed vector space and uses a point-wise distance $d(\vf_i, \vf_j)$ to measure the di-similarity of two items.
In~\cite{contrastive,jing2020self}, $\ell_2$ distance $d(\vf_i,\, \vf_j)=|\vf_i-\vf_j|^2_2$ is used as the item-wise dis-similarity measurement.

Built upon the item-wise measurement $d(\vf_i,\, \vf_j)$, we propose an item-to-set similarity metric $D(S,\, \vf)$, which measures how dis-similar an item $\vf$ is to a set of items $S=\{\vf_1, \cdots, \vf_K\}$. 
The item-to-set metric aims to predict how  similar a outfit candidate is to a set of user selfies for personalized fashion recommendation.
In the following, 
we will first discuss the weakness of two item-to-set metric definitions, then propose our improved item-to-set metric.

We first consider an \emph{averaged} item-to-set distance that computes the averaged distances between all items in a set $S$ and an item $f$, specifically:
\begin{align}
\label{eq:average}
\begin{aligned}
    D_{avg}(S,\, \vf)&=\frac{1}{K}\sum_{i=1}^{K}d(\vf_i,\, \vf)\\
            &=d\left(\frac{1}{K}\sum_{i=1}^{K}\vf_i,\, \vf\right).
\end{aligned}
\end{align}
We note that the averaged distance is equivalent to first averaging all features in the set, then compute the item-wise distance, assuming that $\ell_2$ distance is used for the item-wise distance $d(\cdot,\, \cdot)$. The feature averaging operation is also proposed in~\cite{snell2017prototypical} for few-shot learning.

Alternatively, a \emph{nearest-neighbor} item-to-set distance computes the nearest distance from all items in the set $S$ to query item $f$:
\begin{align}
\label{eq:min}
\begin{aligned}
    D_{NN}(S,\, \vf)&=\min_{i=1}^{K}d(\vf_i,\, \vf)\\
            &=d(\vf_{i^{*}},\, \vf),
\end{aligned}
\end{align}
where $i^* = \arg \min_{i=1}^{K} d(\vf_i,\, \vf)$. It can be seen as a weighted averaged distance, where we assign weight $1$ to item $\vf_{i^*}$ and $0$ to the resting items.

Both of these item-to-set metrics have drawbacks.
First, as the averaged distance performs feature averaging, it cannot properly captures the similarity when $S$ contains items in multiple fashion styles. 
Second, although the nearest-neighbor distance is more adaptive to the multiple fashion style case, the minimum operation is susceptible to outliers and noise. Moreover, it only updates features for the closest items during training. As a result, training with the nearest-neighbor metric could hardly converge.


To design a metric that better captures the multiple interests of a user while facilitating robust training, we propose a generalized item-to-set distance. Specifically, given a set $S$ and a query $f$, we propose to first assign an importance weight $w_i$ to each item $\vf_i \in S$ before feature averaging and distance computation.
The importance weight is computed using an importance estimator $w_i=K(\vf_i; \vf, S)$. Such a {\emph{item-to-set distance}} is defined by:
\begin{align}
\label{eq:weighted}
\begin{aligned}
    D(S, \vf)&= d\left(\sum_{i=1}^{K} \alpha_i\vf_i,\, \vf\right),\\
    \alpha_i &= \frac{\exp(w_i)}{\sum_j \exp(w_j)}.
\end{aligned}
\end{align}
Our formulation is a generalized form of Eq.~\ref{eq:average} and Eq.~\ref{eq:min}. To understand that, note that with $K(\vf_i; \vf, S)$ being a constant value or $-\infty \times d(\vf_i,\, \vf)$, our formulation recovers Eq.~\ref{eq:average} and Eq.~\ref{eq:min}.
However, our importance weights are generated using a learnable function$K(\vf_i; \vf, S)$, which allows our metric to explore a better weight assignment strategy.
Next, we will elaborate on the design of the importance estimator.

\subsubsection{Importance Estimation}
First, we consider {\emph {neighboring importance weight}}:
\begin{equation}
\label{eq:u}
    u(\vf_i; \vf) = - \gamma d(\vf_i,\, \vf),
\end{equation}
where $\gamma$ is a non-negtive and learnable parameter.
The neighboring importance mimics a nearest-neighboring operation in the sense that it assigns more weights to $\vf_i$ that are closer to $\vf$ to capture the multiple interests of a user.
However, unlike nearest neighbor that ignores all non minimal-distance items, the above equation can update all item features during learning for more robust training.
In addition, $\gamma$ is learned from data to balance the trade-off between utilizing all items or only the nearest item.

Due to incorrect parsing, missing modalities or typos, noise and outliers in the set $S$ are inevitable.
To reduce the influences of noise and outliers when computing the distance, we further consider an {\emph {intra-set importance weight}}:
\begin{equation}
\label{eq:v}
    v(\vf_i; S) = \MLP_v\left(\left[\vf_i, stat(S)\right]\right),
\end{equation}
where $\MLP_v$  outputs a scalar from an input vector, and $stat(S)$ is a vector that captures the statistics of the set $S$ along all feature dimensionalities\footnote{We employ mean, standard derivation, min and max functions to compute the statistics along all feature dimensions, then concatenate the extracted statistics into a vector.}. 
In this way, we compare each item $\vf_i$ with the set $S$ to eliminate the outliers from the sets.

Our overall importance weights are generated using a linear combination of Eq.~\ref{eq:u} and Eq.~\ref{eq:v} as follows:
\begin{align}
\label{eq:u+v}
\begin{aligned}
K(\vf_i; \vf, S)&=u(\vf_i; \vf) + v(\vf_i; S).
\end{aligned}
\end{align}

\subsubsection{User-specific Metric Space}
As different users may focus on different aspects of fashion items, the item-to-set metric itself should be user-specific.
For instance, for the minimalist fashion style users, the item-to-set distance should be more sensitive to the amount of colors that are used. However, for users of the artsy style, the item-to-set distance should focus more on unusual prints and the complexity of accessories. 



To extend our similarity metric in Eq.~\ref{eq:weighted} to a user-specific metric, we perform a user-specific space transformation before the distance computation. In particular, given the set $S$, we compute a scaling vector $\vt(S)$ which indicates the scaling factor at each feature dimension:
\begin{align}
\begin{aligned}
\vt(S)=\softmax\left( \MLP_t(stat(S))\right).
\end{aligned}
\end{align}
One could also apply the sigmoid function instead of the softmax function. However, we found that the softmax function slightly boosts the recommendation accuracy because it ensures that all weights sum to $1$.

Using the space transformation, we can extend the item-to-set metric of Eq.~\ref{eq:weighted} to a set-specific metric. Specifically, we define a {\bf{u}}ser-{\bf{s}}pecific item-to-set metric:
\begin{align}
\label{eq:weighted2}
\begin{aligned}
    D_{us}(S,\, \vf)&= d(\vt(S) \odot \left(\sum_{i=1}^{K} \alpha_i\vf_i\right),\, \vt(S) \odot \vf),\\
\end{aligned}
\end{align}
where $\odot$ represents vector elementwise multiplication. Eq.~\ref{eq:weighted2} filters out the feature dimensions that a user focuses less on before the distance computation. This procedure helps the recommendation system to be more user-specific.

\vspace{-1.5mm}
\subsection{Learning Objectives}
\label{sec:loss}
Metric learning generally serves to reduce the distances of positive pairs and enlarge the distances of negative pairs. To adapt this principle for item-to-set metric learning, we sample the item set $S$ of size $K$ from a random user, then generate a positive item $f^{+}$ from the same user and $m$ negative items ${\{f^{-}_j\}}_{j=1}^{m}$ from the other users. The item-to-set metric learning is cast as an $(m+1)$-way classification problem, which aims to classify the positive samples from all negative samples. In particular, we minimize the negative log-likelihood as follows:
\begin{align}
\label{eq:objective}
\begin{aligned}
    \mathcal{L}(S,f^{+},{\{f^{-}_j\}}_{j=1}^{m}) = -\log \frac{\exp(-D(S,f^{+}))}{\exp(-D(S,f^{+}))+\sum_{j=1}^{m} \exp(-D(S,f_j^{-}))}.
\end{aligned}
\end{align}
We employ the user-specific item-to-set distance (Eq.~\ref{eq:weighted2}) as the item-to-set distance function. In testing stage, given a user selfie post set $S$, we recommend items that are close to the set with the learned item-to-set distance.

\vspace{-3mm}
\section{Experiments}
To evaluate the effectiveness and performance of our approach, we conduct experiments using the dataset collected in Sec.~\ref{sec:data}.
To this end, we reserve the latest activities of all the 2,293 users as a outfit candidate pool for recommendation. This candidate pool is used to evaluate the recommendation performance.
Afterwards, we randomly split the 2,293 users into a training set, which contains  1834 users, and a test set, which contains the remaining users.
Such a data split ensures that training set, test set and outfit pool are disjoint for fair evaluation.
We perform model training on the training set and evaluate the fashion recommendation results on the test set. 

\vspace{-3mm}
\subsection{Implementation}
Our metric learning framework is implemented using PyTorch~\cite{NEURIPS2019_9015}.
Although our item-to-set metric computes the distances between set items and multiple query items, our implementation utilizes standard matrix operations to support efficient batched training.
We set the initial learning rate to $0.001$ and decay it by a factor of $0.2$ after every $300$ epochs and optimize the weights via SGD with a momentum of $0.95$.
The batchsize is set to $32$ in our experiments while the number of negative items to set to $m=50$ for most of our experiments. We also evaluate the influence of $m$ with more experiments.
For evaluation, we randomly select $n=10$ user selfie posts as input from each user for recommendation.
We repeat the sampling process for $50$ times and report the averaged recommendation performance. The random seed for sampling is fixed during evaluation.
We also test other input sizes with more experiments.

\vspace{-3mm}
\subsection{Quantitative Evaluation}
The quantitative performance for fashion recommendation is evaluated with top-k recall at $k=1,10,25$.
In Table~\ref{tab:baseline}, we compare our method with two item-wise metric learning schemes \emph{triplet+avg} and \emph{triplet+avg}, which learn selfie post embedding and respectively computes item-to-set distance with average distance or nearest distance for recommendation.
In addition, a recent few-shot-learning method~\cite{li2019revisiting} proposed the image-to-class measure that measures similarity from an image to a set of queries. We incorporate the image-to-class measure (with 3-NN) as item-to-set metric and train a baseline~\emph{DN4} using triplet loss for recommendation. Our model and the comparative methods are trained with the same sampling strategy and train setting for fair comparisons.
From the table, our approach based on item-to-set metric achieves 230 times improvement over random guess in terms of recall@1 and shows substantiate advantages over the comparative methods.
We also perform extensive comparative experiments to study effectiveness our design, which are elaborated as follows:

\begin{table}[]
	\caption{
	The fashion recommendation performance in comparisons to different methods. Performances are measured in recall scores at 1, 10 and 25 respectively.}
    \vspace{-3mm}
	\centering
	\scalebox{1.0}{
	\begin{tabular}{ l|c|c|c}
		\toprule
		\hline
		Methods       &	Recall@1 & Recall@10 & Recall@25\\
		\hline
		random guess& 0.0004&    0.0043&    0.0108\\
		triplet+NN      & 0.0141& 0.0589&0.1037\\
		triplet+avg      & 0.0144& 0.0650& 0.1085\\
		DN4~\cite{li2019revisiting}& 0.0163&0.0741& 0.1283\\
		ours & \best{0.1005}&    \best{0.2420}&    \best{0.3336}\\
		\hline
		\bottomrule
	\end{tabular}
	}
	\label{tab:baseline}
\end{table}

\vspace{1mm}
{\noindent \bf Multimodalities and fusion.} To study the benefit of using multi-modal features, we train our fashion recommendation system with different combinations of modalities and fusion methods, and report their performances. 
We also evaluate the different implementation of the embedding module, i.e. hashtag and title modalities with or without using attentive feature averaging (denoted by \emph{w/} or \emph{w/o att}).
We compare our cross-modal gated fusion with fusion by concatenation (denoted by \emph{w/} or \emph{w/o cross}).
From Table~\ref{tab:feature}, we observe that:
i) hashtag is the most informative modality as it often contains fashion style descriptions of outfits,
ii)  attentive feature averaging can improve the representative power of hashtag and title modalities,
iii) multi-modality fusion improves the recommendation performance,
iv)  cross-modal gated fusion can improve the recommendation performance.

\begin{table}[http!]
\caption{The impacts of modalities and fusion schemes on the performance of fashion recommendation. Performances are measured in recall at 1, 10 and 25, respectively.}
\vspace{-3mm}
\centering
\scalebox{1.0}{
\begin{tabular}{l|c|c|c}
\toprule
\hline
Baselines       &	Recall@1 & Recall@10 & Recall@25\\
\hline
Image (I)               & 0.0236&    0.1003&    \best{0.1666}\\
Hashtag (H) \emph{w/o att}& 0.0601&    0.1128&    0.1485\\
Hashtag (H) \emph{w/ att}& \best{0.0738}&    \best{0.1293}&    0.1630\\
Title (T) \emph{w/o att}& 0.0183&    0.0693&    0.0988\\
Title (T) \emph{w/ att}& 0.0268&    0.0687&    0.1012\\
\hline
T+H \emph{w/o att}, \emph{w/o cross} & 0.0766&    0.1572&    0.2097\\
T+H \emph{w/ att} \emph{w/o cross} & 0.0841&    0.1730&    0.2178\\
T+H \emph{w/ att} \emph{w/ cross} & \best{0.0901}&    \best{0.1736}&    \best{0.2219}\\
\hline
I+T+H \emph{w/o att}, \emph{w/o cross} & 0.0857&    0.2185&    0.3040\\
I+T+H \emph{w/ att} \emph{w/o cross} & 0.0914&    0.2289 &   0.3118\\
I+T+H \emph{w/ att} \emph{w/ cross} & \best{0.1055}& \best{0.2420} & \best{0.3336}\\
\hline
\bottomrule
\end{tabular}
}
\label{tab:feature}
\end{table}

\begin{figure}[http!]
  \centering
  \includegraphics[width=1.\linewidth]{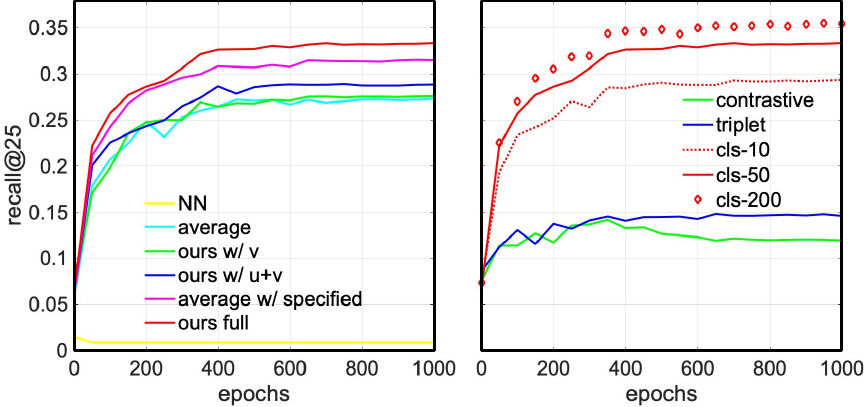}
  \caption{The convergence curves of recommendation models with different item-to-set metric schemes. Recommendation performances are measured using recall scores at 25.}
  \label{fig:new}
\end{figure}

{\noindent \bf Metric designs.}  
In Table~\ref{tab:metric} and Fig.~\ref{fig:new} left, we evaluate  different variants of our proposed item-to-set metrics, and show their convergence curves:
i) \emph{NN}, denoting the nearest-neighbor item-to-set distance defined by Eq.~\ref{eq:min},
ii) \emph{average}, denoting the averaged item-to-set distance defined by Eq.~\ref{eq:average},
iii) \emph{ours w/ v}, denoting a metric that uses importance weights but only applies intra-set importance as described in Eq.~\ref{eq:v},  
iv) \emph{ours w/ u+v}, denoting a metric that applies both intra-set importance and neighboring importance as described in Eq.~\ref{eq:u+v},
v) \emph{average w/ specified}, denoting a metric that applies averaged distance but also applies the user-specific formulation in Eq.~\ref{eq:weighted2}
v) \emph{ours full}, denoting our full metric.
From Table~\ref{tab:metric} and Fig.~\ref{fig:new} left we observe that: 
i) \emph{NN} cannot converge as it is susceptible to embedding noises while the performance of \emph{average} is the second lowest
ii) with our formulation, intra-set importance \emph{ours w/ v} and neighboring importance \emph{ours w/ u+v} can improve over \emph{average},
iii) the user-specific formulation can improve performance upon both \emph{average} and \emph{ours w/ u+v}. Notably, our full metric is able to double the recall@1 in comparison with \emph{average}.

\begin{table}[]
	\caption{
	The fashion recommendation performance for different item-to-set metric schemes. Performances are measured in recall scores at 1, 10 and 25 respectively.}
    \vspace{-3mm}
	\centering
	\scalebox{1.0}{
	\begin{tabular}{ l|c|c|c}
		\toprule
		\hline
		Methods       &	Recall@1 & Recall@10 & Recall@25\\
		\hline
		NN      & 0.0000&    0.0044&    0.0087\\
		average & 0.0587 &   0.1872&    0.2731\\
		ours \emph{w/ v} & 0.0631&    0.1961&    0.2759\\
		ours \emph{w/ u+v} & 0.0794&    0.2253&    0.3150\\
		average \emph{w/ specified} & 0.0805&    0.2038&    0.2885\\
		ours full & \best{0.1005}&    \best{0.2420}&    \best{0.3336}\\
		\hline
		\bottomrule
	\end{tabular}
	}
	\label{tab:metric}
\end{table}

{\noindent \bf Learning objectives.} In Table~\ref{tab:objective} and Fig.~\ref{fig:new} right, we further evaluate other training objectives other than our proposed objective. In particular, we test the \emph{contrastive} loss~\cite{contrastive} and \emph{triplet} loss~\cite{triplet} in our item-to-set setting. We also vary the negative sample size $m$ from Eq.~\ref{eq:objective} to analyze the impact on it (different results are denoted by \emph{cls-$m$}). We observes that for our task, contrastive and triplet objectives are not as effective as the classification objective. In addition, increasing the negative sample size $m$ will improve the recommendation performance.

\begin{table}[]
	\caption{The fashion recommendation performance for different training objectives. Performances are measured in recall scores at 1, 10 and 25 respectively.}
    \vspace{-3mm}
	\centering
	\scalebox{1.0}{
	\begin{tabular}{ l|c|c|c}
		\toprule
		\hline
		Methods       &	Recall@1 & Recall@10 & Recall@25\\
		\hline
		random guess& 0.0004&    0.0043&    0.0108\\
		contrastive  & 0.0231 &   0.0738  &  0.1193\\
		triplet         & 0.0162 &   0.0831 &   0.1460 \\
		cls-10 & 0.0756&   0.2061&    0.2937\\
		cls-50  & 0.1005&    0.2420&    0.3336\\
		cls-200  & \best{0.1210}&    \best{0.2627}&    \best{0.3545}\\
		\hline
		\bottomrule
	\end{tabular}
	}
	\label{tab:objective}
\end{table}


\begin{figure}[]
  \centering
  \includegraphics[width=0.9\linewidth]{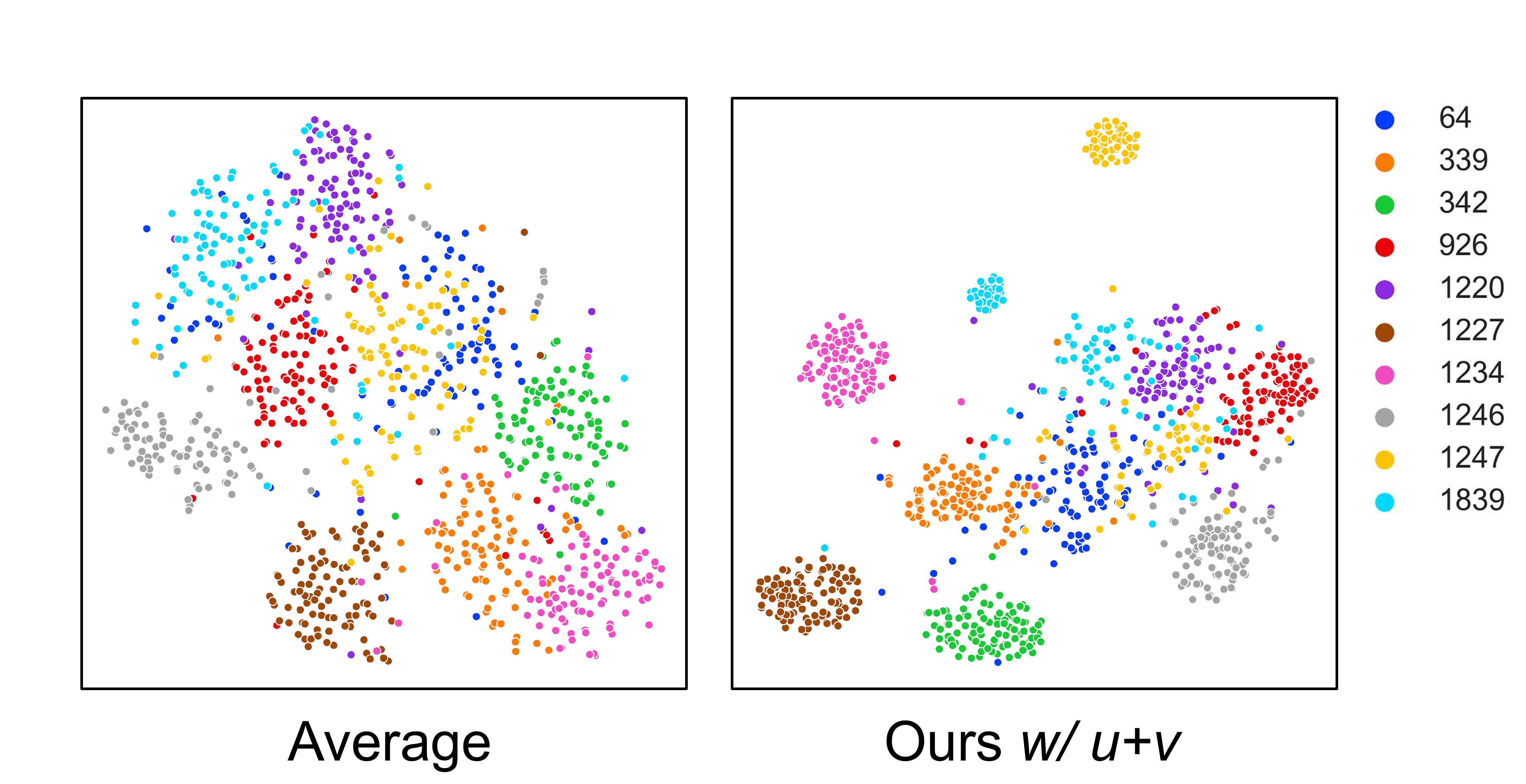}
  \caption{The user selfie embedding learned with the averaged item-to-set metric (left) and our metric with importance weights. Our metric represents a user (e.g. colored by yellow or cyan) with multiple clusters, which effectively captures the multiple fashion interests of a user.}
  \label{fig:tsne}
\end{figure}

{\noindent \bf Influence of set size.}
We also study the influence of set size. Specifically, we vary the input activities size of users $n$ and observe the recommendation performances. From Tab.~\ref{tab:set_size}, lager set size improves the recommendation performance as we expected.


\begin{table}[]
	\caption{The fashion recommendation performance with different numbers of input selfie posts $n$. Performances are measured in recall scores at 1, 10 and 25 respectively.}
    \vspace{-3mm}
	\centering
	\scalebox{1.0}{
	\begin{tabular}{ l|c|c|c|c|c}
		\toprule
		\hline
	     & $n=3$ & $n=5$ & $n=10$ & $n=15$ & $n=20$\\
	    \hline
        Recall@1 & 0.0735 & 0.0897 & 0.1005 &  0.1144& \best{0.1230}\\
        Recall@10 & 0.1786 & 0.2146 & 0.2420 & 0.2611 & \best{0.2798}\\
        Recall@25 & 0.2492 & 0.2919 & 0.3336 & 0.3541 & \best{0.3813}\\
		\hline
		\bottomrule
	\end{tabular}
	}
	\label{tab:set_size}
\end{table}

\begin{figure}[]
  \centering
  \includegraphics[width= \linewidth]{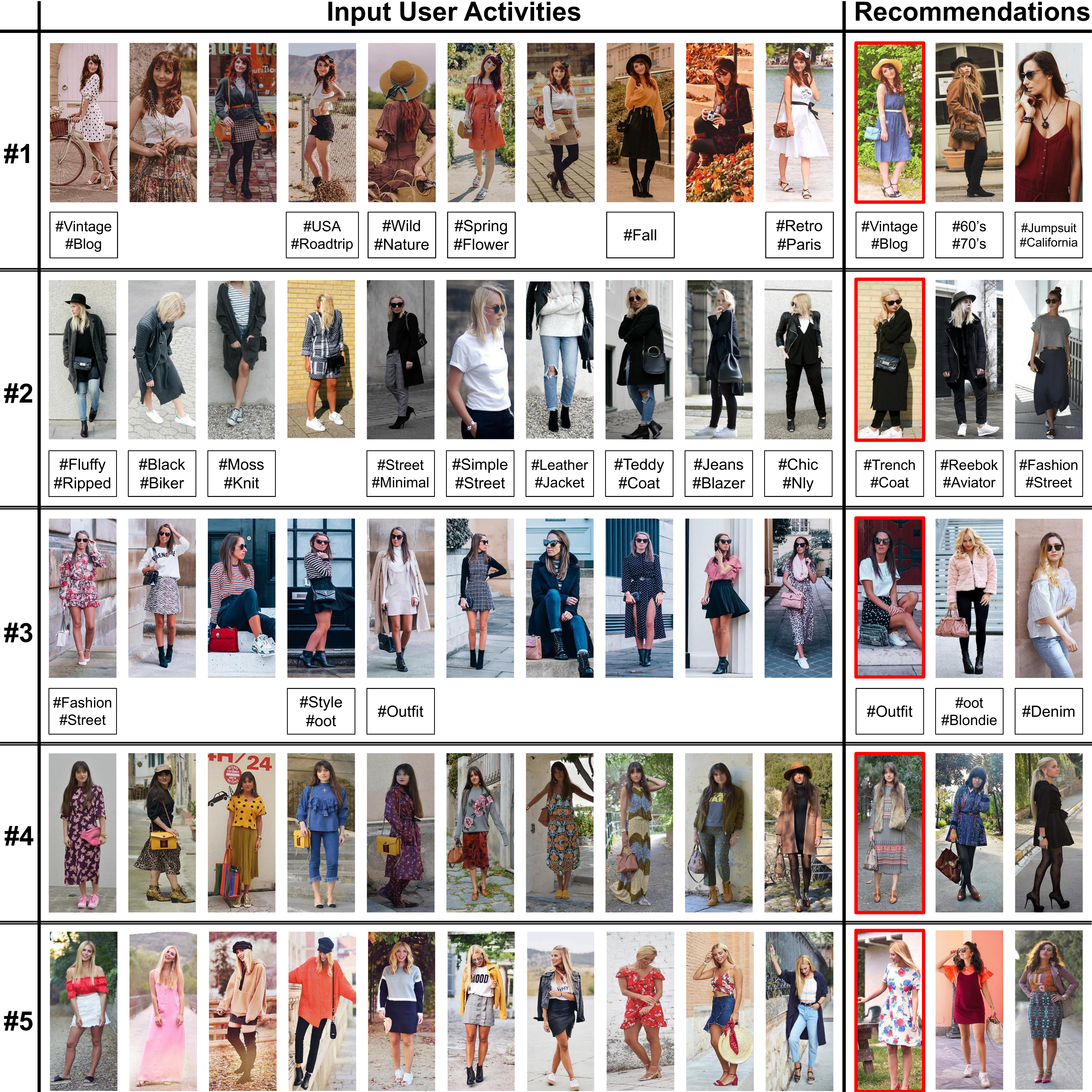}
  \caption{Based on posts of users (Left Columns), outfits are recommended and top 3 are shown (Right columns). A red box represents when our recommendation system correctly predicts the outfit of users.
  18.9\% of the test users are given the correct outfits with our top-3 recommendation (87 out of 459).
  This shows that our model can makes good personalized recommendations based on the fashion taste of users.}
  \label{fig:result2}
\end{figure}

\subsection{Qualitative Results}
\subsubsection{Recommendation Results}
For qualitative evaluation, we depict the top-$3$ recommendation results of our model on the test set. The top-3 recall of our model is 18.9\%, meaning that it gives correct outfits recommendation to 18.9\% of the test users (87 out of 459).
As shown in Fig.~\ref{fig:result2} (right), our recommendation system is able to recommend style coherent outfits to individual users (a red box indicates a correct prediction on the test set).
For instance,
user $1$ prefers vintage and country styles and the combination of dresses and skirts with heels or boots.
From posts \#3, \#7, \#8 and \#9, we learn that for the fall season, she prefers neutral colors such as black and brown with boots.
She also prefers boater or beret hat as embellishments (Posts \#5 and \#9).
For user $1$, our first recommendation (k=1) correctly predict the outfit of the user, which consists of a vintage style blue dress with a textured straw hat.
It also successfully recommends her preferred outfit style for the fall season (k=2), which consists of a brown jacket with black boots and a beret hat. 
Our third recommendation is a U.S. country style burgundy jumpsuit (k=3), which also matches the dressing tastes of user $1$.

Our recommendations to users $2$ and $5$ further substantiate the personalized recommendation capacity of our model.
For instance, user $2$ prefers street style outfits such as jackets and jeans in neutral colors (white, gray and black), white sport shoes or black boots.
For user $2$, our model recommends black jackets with white sport shoes (k=1,2) which match her preferred items. 
It also recommends a simple street styled outfit with gray t-shirt and dart gray maxi skirt (k=3), which matches the dressing style of user $2$.
User $5$ prefers warm colors and floral patterns for her outfits. Our model recommends a white dress with contrast coloring floral patterns (k=1), which is her own outfit. It also recommends a terse dress with a combination of warm colors (k=2), which also matches the color preference of the user. Readers are referred to the supplementary material for more visualization of our recommendation results.

\vspace{-1mm}
\subsubsection{Effects of Importance Weights}
It is useful to understand the role that the importance weights play. In Fig.~\ref{fig:tsne}, we visualize the embedding of the average item-to-set baseline, as well as the embedding learned using our importance weights (the baseline \emph{ours u+v} in Table~\ref{tab:metric}).
It can be observed that \emph{average} tends to represent user posts with loosely cluster points. In contrast, \emph{ours u+v} represents user posts with multiple tightly clusters points, which effectively represents the multiple fashion interests of a user for more accurate and flexible recommendation.


\vspace{-3mm}
\section{Conclusion}
In this work, we study the problem of personalized fashion from personal social media data. We present a item-to-set metric learning framework that learns the similarity between user posts and fashion outfits. 
To account for the diversity of fashion interests of users, we propose neighboring importance.
To reduce the influence of noise and outliers in a set, we propose intra-set importance.
The combination of the two terms serves to dynamically assign weights for adaptive item-to-set similarity measurement. 
We further propose user-specific space transformation that learns user-specific metrics for more personalized recommendation.
To extract features from user activities and outfits, we propose a multi-modality feature extraction module for cross-modality fusion. 
We collect a real-world social media dataset to access the performance of fashion recommendation.
The effectiveness of our framework is shown through extensive experiments and analysis.


\bibliographystyle{ACM-Reference-Format}
\bibliography{egbib}

\end{document}